# SUMMARY ON THE ISCSLP 2022 CHINESE-ENGLISH CODE-SWITCHING ASR CHALLENGE

*Shuhao Deng[1], Chengfei Li[1,*], Jinfeng Bai[1], Qingqing Zhang[2], Wei-Qiang Zhang[3], Runyan Yang[4], Gaofeng Cheng[4], Pengyuan Zhang[4], Yonghong Yan[4]*

[1]TAL Education Group, Beijing, China  [2]Magic Data  [3]Tsinghua University  [4]Institute of Acoustics, Chinese Academy of Sciences

{dengshuhao1, lichengfei, baijinfeng1}@tal.com, zhangqingqing@magicdatatech.com, wqzhang@tsinghua.edu.cn, {yangrunyan, chenggaofeng, zhangpengyuan, yanyonghong}@hccl.ioa.ac.cn

## Abstract

Code-switching automatic speech recognition becomes one of the most challenging and the most valuable scenarios of automatic speech recognition, due to the code-switching phenomenon between multilingual language and the frequent occurrence of code-switching phenomenon in daily life. The ISCSLP 2022 Chinese-English Code-Switching Automatic Speech Recognition (CSASR) Challenge aims to promote the development of code-switching automatic speech recognition. The ISCSLP 2022 CSASR challenge provided two training sets, TAL_CSASR corpus and MagicData-RAMC corpus, a development and a test set for participants, which are used for CSASR model training and evaluation. Along with the challenge, we also provide the baseline system performance for reference. As a result, more than 40 teams participated in this challenge, and the winner team achieved 16.70% Mixture Error Rate (MER) performance on the test set and has achieved 9.8% MER absolute improvement compared with the baseline system. In this paper, we will describe the datasets, the associated baselines system and the requirements, and summarize the CSASR challenge results and major techniques and tricks used in the submitted systems.

## 1. Introduction

In recent years, with the rapid development of globalization, people often mix other languages in the Chinese context in their daily communication, which is called code-switching phenomenon. The code-switching phenomenon, especially the mixed expression in Chinese and English, is very common in application scenarios, such as workplace and English teaching scene. In some multilingual societies, such as Malaysia and Singapore, code-switching phenomenon occurs almost in everyday conversations. Generally speaking, in code-switching phenomenon, the speakers often alternate between different languages within an utterance or between consecutive utterances. As a result, Chinese-English code-switching can be divided into two forms of expression: one is inter-sentence mixing, and another is intra-sentence mixing. For example, "Oh, My God, 我的电脑死机了", which means "Oh, My God, my computer crashed" in English, is a typical is inter-sentence mixing phenomenon. In this sentence, the first part of the sentence is composed of English sentences, and the second part of the sentence is composed of Chinese sentences. And "我要上 Coursera 学习 Machine Learning 课程", which means "I'm going to Coursera to learn Machine Learning course" in English, is a typical intra-sentence mixing phenomenon. In this sentence, we can think that the main body of the sentence is Chinese, and some Chinese characters have been replaced by English words. Generally speaking, such two expressions are considered correct and grammatical.

Because of the existence of code-switching phenomenon and the complexity of multilingual languages, CSASR is usually more difficult than monolingual language speech recognition. In addition, the shortage of corresponding code-switching corpus also affects the development of CSASR to a large extent. With the popularization of automatic speech recognition equipment, how to make machines better learn and understand the complex and changeable language of human beings has become a huge challenge faced by many machine learning researchers and practitioners. The matching degree and richness of data is one of the most important factors to promote the performance of automatic speech recognition system. Only by training high-quality structured data can the machine understand the speaker's intentions and instructions and make more accurate responses. To promote the development of CSASR [1], the ISCSLP 2022 Chinese-English Code-Switching Automatic Speech Recognition (CSASR) Challenge was held, and this CSASR challenge hopes to work with enterprises, institutions and universities to jointly explore innovative development in the artificial intelligence field of Chinese-English code-switching dialogue. The CSASR challenge provided two training sets, TAL_CSASR corpus [2] and MagicData-RAMC corpus [3], a development and a test set for participants, which are used for CSASR model training and evaluation. Along with the challenge, we also provide the baseline system performance for reference. As a result, more than 40 teams participated in this challenge, and the winner team achieved 16.70% MER performance on the test set and has achieved 9.8% absolute MER improvement compared with the baseline system.

This paper summarizes the outcomes of CSASR challenge. Specifically, we give a brief overview on Chinese-English code-switching ASR in Section 2. Section 3 reviews the datasets, the associated baseline system and the requirements. Sections 4 discuss the outcome of the CSASR challenge with

---

major techniques and tricks used in submitted systems. Section 5 concludes the paper.

## 2. Related Works

Due to the development of machine learning and neural network, monolingual speech recognition, such as Chinese speech recognition and English speech recognition, has achieved great success. In addition, another main reason for the success of monolingual speech recognition is that a large number of labeled data are used to train ASR model. In recent years, Chinese-English code-switching automatic speech recognition have attracted increasing attentions. And CSASR also benefits from the development of machine learning and neural networks, as well as the accumulation of corresponding code-switching labeled data. In this section, we will briefly review the development of machine learning and neural networks related to CSASR and some code-switching corpus.

### 2.1. Networks

In ASR community, the sequence-to-sequence acoustic modeling has attracted great attention. In the past ten years, deep neural network (DNN) [4] has gradually replaced Gaussian mixture model (GMM) for acoustic modeling, and the hybrid ASR system composed of acoustic model, language model and lexicon model has achieved convincing performance. Recently, end-to-end (E2E) ASR models such as Connectionist Temporal Classification (CTC) [5], the recurrent neural network transducer (RNN-T) [6], transformer [7] or conformer transducer [8], attention-based encoder-decoder models [9] have gained popularity and achieved state-of-the-art performance in accuracy and latency. In contrast to conventional hybrid ASR systems, they joint learn acoustic and language modeling in a single neural network that is E2E trained from labeled data. For example, the recently proposed Transformer-based end-to-end ASR architectures use deeper encoder-decoder architecture with feedforward layers and multi-head attention for sequence modeling, and comes with the advantages of parallel computation and capturing long-contexts without recurrence.

To handle code-switching speech, there have been many studies in acoustic modeling [10], language modeling [11, 12], and ASR systems [13-15]. And Chinese-English CSASR is typically a low-resource task due to the scarce acoustic and text resources that contain code-switching. In CSASR, to solve the data scarcity problem during acoustic modeling, one solution is to simply merge two different monolingual datasets into one code-switching dataset for the training of CSASR model, another effective solution is to adapt two well-trained high-resource language acoustic model to the target low-resource domain using transfer learning. These solutions can alleviate the problem of data scarcity to some extent, but the performance of CSASR model will reach the bottleneck and cannot be further improved because its training data does not fully match the situation in the actual application scenario.

### 2.2. Code-switching corpus

For CSASR, A large number of well labeled code-switching data that can be used to train the CSASR model is critical. To promote the development of CSASR, some companies and organizations released some code-switching data for academic research. To our best knowledge, there are several code-switching datasets which is freely available.

SEAME corpus [16] provides about 112 hours Mandarin-English code-switching conversational speech data, which is collected from residents of Malaysia and Singapore. English-Mandarin codeswitching speech recognition contest sponsored by DataTang [17] in China released two training datasets, namely, 200 hours of English-Mandarin code-switching data, and 500 hours of Mandarin data of which there are about 15 hours of similar code-switching data. And this means that there are 215 hours of code-switching data available for academic research. TAL_CSASR corpus [2], released by TAL education group, contain more than 100 speakers and over 580 hours of Mandarin-English code-switching speech data. TAL_CSASR corpus is one of the most representative scenes of code-switching—teacher English teaching scene, and it is also the known largest open-source Mandarin-English code-switching corpus in the world.

## 3. Datasets, Baseline System, and Requirements

The ISCSLP 2022 CSASR Challenge Organizing Committee provided the corresponding datasets for participants to use in training and evaluating CSASR models. Also, the Organizing Committee provide the baseline system performance for reference. And participants must train and evaluate the CSASR model under the requirements of the Organizing Committee. The evaluation system is used to evaluate the CSASR model performance of participants and determine the final ranking. In the follow, we will describe the details.

### 3.1. Datasets

The Organizing Committee provided two training sets, a development and a test set for participants, which are used for CSASR model training and evaluation. The training sets consists of the TAL_CSASR corpus and the MagicData-RAMC corpus. The TAL_CSASR corpus contain more than 100 speakers and over 580 hours of Mandarin-English code-switching speech data. TAL_CSASR corpus comes from online one-to-one teaching scene, in which teachers and students come from different regions of China, and each audio contains only one speaker. The MagicData-RAMC corpus includes 351 groups of multiple rounds of Mandarin dialogue, with a total duration of 180 hours. The annotation information of each group of dialogue includes the transcript, the voice activity timestamp, the speaker information, the recording information and the topic information. The speaker information includes gender, age and region, and the recording information includes environment and equipment. The duration of development set is about 6.8 hours, including 14 speakers. And the duration of test set is about 2.7 hours, including 28 speakers. The development set and test set here belong to the same source data. Two noise datasets can be used for data augmentation, namely MUSAN (openslr17) [18] and RIRNoise (openslr 28) [19]. The CSASR model must be trained using only the provided training datasets, combining with the provided two noise datasets. Participants can use the development set to evaluate their CSASR model, but the test set was used to rank final performance.

### 3.2. Baseline System

The baseline system of the ISCSLP 2022 Code-Switching ASR Challenge is developed using ETEH platform[1] [20], and the Transformer is used as the model architecture. A 2-layer convolutional neural network (CNN) was employed as the front-end. Each CNN layer has 320 filters, each of which has 3x3 kernel size with 2x2 stride. The self-attention encoder and decoder are 17-layer and 6-layer, respectively. All sub-layers, as well as embedding layers, produce outputs of dimension 320. In the multi-head attention networks, the head number is 8. The inner dimension of position-wise feed-forward networks is 2048. All the ASR models are trained with batch size 512, using Adam algorithm with gradient clipping norm 5, warm-up of 25000 steps, and Noam learning rate decay scheme. The MagicData-RAMC Trainset and TAL_CSASR Trainset were combined as the training dataset for training the baseline CSASR model. And 83-dimensional features, which include 80-dimensional filter banks and 3-dimensional pitch features, are used as input acoustic features. Features are extracted with a 25ms Hamming window, shifted every 10ms. Cepstral mean and variance normalization (CMVN) is not applied to the acoustic features. The baseline CSASR model predicts subword units based on byte pair encoding (BPE) for English and Chinese characters for Mandarin as output targets. There are 5276 output targets in total, of which 1007 are English subword units (including special symbols) and 4269 are Chinese characters. The baseline CSASR model was trained for 30 epochs for code-switching automatic speech recognition. The open source toolkit, Sclite, is employed to evaluate performance of the baseline CSASR model and the submitted CSASR model. The Mixture Error Rate (MER) is adopted as the evaluation criterion, which calculates the character error rate (CER) for Chinese and the word error rate (WER) for English. As a result, the baseline CSASR model achieve 29.2% MER performance on the development set and 26.5% MER performance on the test set.

---

[1] https://github.com/SpeechClub/ETEH

### 3.3. Requirements

The Organizing Committee has set corresponding rules to ensure the fairness of the ISCSLP 2022 CSASR Challenge. Specifically, the Organizing Committee has restricted the scope of training dataset for CSASR model training. This means that participants cannot use extra training data, except for two ASR corpus, TAL_CSASR corpus and MagicData-RAMC corpus, and two noise datasets using for data augmentation, MUSAN (openslr17) and RIRNoise (openslr 28). The test sets cannot be used in any form, such as model training and model fine-tuning. Multi-model fusion is allowed, but multi-model fusion with the same structure is not encouraged. All CSASR models must be trained on the provided datasets, and this means the pre-training model also does not allow the use of extra datasets, including unlabeled datasets.

## 4. Summary on Submitted System

The ISCSLP 2022 CSASR Challenge has received more than 40 teams' registration and participation, which cover many companies, universities, academic research institutions and individual researchers in the field of automatic speech recognition. These teams employed different model structures and optimization strategies or tricks to improve the performance of the CSASR model, and achieved considerable improvement. Table 1 summarizes the model structure and optimization strategies or tricks used by the top six teams. As the Table 1 shows, the winner of the ISCSLP 2022 CSASR Challenge goes to team "conv". In the follow, this paper will analyze and discuss the system submitted by each team from four aspects: model structure, data augmentation and post-processing and other tricks.

### 4.1. Model Structure

As the Table 1 shows, the Conformer structure is the most commonly used acoustic model, and five teams used the Conformer structure in total, which also proves the success of the Conformer structure in the field of automatic speech recognition.

The team "conv" uses 15 comformer block as encoder and 6 transformer block as decoder, which means team "conv" has a deeper encoder in acoustic model structure than other teams. It is worth noting that team "What_to_eat" not only uses the conventional Conformer structure, but also uses the bi-encoder [21] structure, language-aware encoder (LAE) [22] structure as well as the Mixture-of-Experts (MoE) [23] structure as the acoustic model. And team "What_to_eat" trains a series of different acoustic model structures, then combines the results of these acoustic models to produce the final results. Unlike other teams, team "tingyin" uses the Squeezeformer [24] structure rather than the conformer structure, and the Squeezeformer structure is a variant of the Conformer structure. According to the report of team "tingyin", under the condition that the model parameters are the same, the Squeezeformer structure can get about 1% MER absolute improvement compared with the Conformer structure on the development set.

### 4.2. Data Augmentation

In order to train a CSASR model with good performance, a large amount of training data is essential. Since the ISCSLP 2022 CSASR Challenge only provides TAL_CSASR corpus and the MagicData-RAMC corpus for training, about 760 hours, as the Table 1 shows, all teams have used data augmentation methods to expand their training sets in varying degrees.

Almost all teams have used common data augmentation methods such as adding Noise and Reverberation, Speed&Volume Perturb, and Spectrum Augmentation. In addition, team "What_to_eat" and team "12321" used Pitch Shift to expand their training set. According to the report of team "What_to_eat", Speed Perturb and Pitch Shift can achieve 0.96% and 0.77% MER absolute improvement respectively on the development set. And according to the report of team "cscscs-asr", Spectrum Augmentation can achieve 0.60% MER absolute improvement on the development set. Also, according to the report of team "SpeechDream", Spectrum Augmentation can achieve 4.28% MER relative improvement on the development set. It is noteworthy that team "What_to_eat" also uses Opus codec to encode and decode the training data to expand the training set, which also can make slighter improvements on the development set. In addition, the team "What_to_eat" also used Text to Speech (TTS) technology to generate corresponding audio for expanding the training set.

### 4.3. Post-processing

As the Table 1 shows, the Post-processing strategies, such as the use of language models (LM) and model fusion technology, also contribute to the performance improvements of the submitted system. Four teams, "conv", "What_to_eat", "cscscs-asr" and "SpeechDream", employed language models to improve the performance of their submitted systems. The team "conv" and the team "cscscs-asr" employed n-gram LM to improve the performance of their submitted systems. N-gram LM can achieve 0.99% and 1.20% MER absolute improvement on the development set, according to the report of the team "conv" and the team "cscscs-asr". The team "SpeechDream" employed RNN-LM to improve the performance of their submitted systems. The RNN-LM can achieve 2.74% MER relative improvement on the development set. The team "What_to_eat" employed

Table 1. The MER performance of top 6 ranking teams and their model structures and optimization strategies or tricks

| Team | Model Structure | | | | Data Augmentation | | | | | Post-processing | Other Tricks | | | MER(%) in test set |
|---|---|---|---|---|---|---|---|---|---|---|---|---|---|---|
| | Transformer | Conformer | Moe | Squeezeformer | Noise & Reverberation | Speed & Volume Perturb | Spectrum Aug | Pitch Shift | TTS & Codec | LM | Model Fusion | Loss Function | Pseudo Text | |
| conv | | ☑ | | | ☑ | ☑ | ☑ | | | ☑ | | | ☑ | 16.70 |
| What_to_eat | | ☑ | ☑ | | | ☑ | ☑ | ☑ | ☑ | ☑ | ☑ | ☑ | | 16.90 |
| tingyin | | | | ☑ | ☑ | ☑ | ☑ | | | | | | | 20.10 |
| cscscs-asr | | ☑ | | | ☑ | ☑ | ☑ | | | ☑ | | ☑ | | 20.90 |
| 12321 | | ☑ | | | ☑ | ☑ | ☑ | ☑ | | | | | | 21.60 |
| SpeechDream | | ☑ | | | | | ☑ | | | ☑ | | | | 23.30 |
| Baseline | ☑ | | | | | | | | | | | | | 26.50 |

Transformer LM and LongContext Transformer LM to improve the performance of their submitted systems. And both Transformer LM and LongContext Transformer LM can achieve different degrees of MER improvement on the development set. In terms of model fusion, the team "What_to_eat" adopted different combinations of acoustic models and language models, and then trained seven CSASR models. And they used the recognizer output voting error reduction (ROVER) [25] to generate the final results of submitted system. And ROVER determines the final CSASR result according to the frequency and confidence of the results obtained by multiple systems.

### 4.4. Other Tricks

In addition to model structure design, data augmentation method and post-processing, there are also other tricks to improve CSASR model performance on the development set, such as loss function design and pseudo text construction.

In terms of loss function design, in addition to using common attention loss and ctc loss, team "What_to_eat" also employed consistency loss [26] combined with the audio generated by TTS technology to improve the performance of the CSASR model. And the team "cscscs-asr" employed the Language Identification (LID) loss as a part of the loss function, and combine the LID loss with attention loss and ctc loss to optimize their CSASR model. According to the report of the team "cscscs-asr", compared with using only attention loss and ctc loss, combining the LID loss with attention loss and ctc loss can achieve 2.2% MER absolute improvement on the development set, which is a considerable performance improvement. In terms of pseudo text construction, team "conv" uses Chinese text to construct pseudo Chinese-English code-switching text for training language model, which is a noteworthy trick. According to the report of the team "conv", constructing pseudo Chinese-English code-switching text to train language model can achieve 0.1% MER absolute improvement on the development set. Moreover, text normalization (TN) techniques are also used to standardize their lexicon. And team "cscscs-asr" use jieba text segmentation tool to expand their text corpus. And all of these ticks can achieve performance improvement on the development set.

## 5. Conclusions

This paper summarized the ISCSLP 2022 Chinese-English Code-Switching Automatic Speech Recognition Challenge and discuss the model structure and optimization strategies or tricks of the submitted system. Due to the existence of code-switching phenomenon and the complexity of multilingual languages, CSASR is usually more difficult than monolingual language speech recognition. In addition, the shortage of corresponding code-switching corpus also affects the development of CSASR to a large extent. The ISCSLP 2022 CSASR Challenge aims to promote the development of code-switching automatic speech recognition, and provide two training sets, TAL_CSASR corpus and MagicData-RAMC corpus, a development and a test set for participants, and provide the baseline system performance for reference. More than 40 teams participated in this challenge, which cover many companies, universities, academic research institutions and individual researchers in the field of automatic speech recognition. Participants used different strategies to optimize the submitted system, such as model structure design, data augmentation, post-processing methods and other tricks. The Conformer structure is the most commonly used acoustic model due to its success in the field of automatic speech recognition. In order to train a CSASR model with good performance, a large amount of training data is essential. Therefore, data augmentation becomes the most popular model optimization sstrategy. In addition, language model is also a popular choice. As a result, the winner team achieved 16.70% MER performance on the test set and has achieved 9.8% MER absolute improvement compared with the baseline system.

## 6. Acknowledsgements

This work was supported by National Key R&D Program of China, under Grant No.2020AAA0104500.